\documentclass{article}
\usepackage[ruled,vlined]{algorithm2e}
\usepackage{spconf,amsmath,graphicx,amssymb,booktabs}

\ninept
\DontPrintSemicolon

\usepackage{enumitem}
\setlist{nosep, leftmargin=14pt}

\usepackage{mwe} 

\usepackage{verbatim}
\usepackage[normalem]{ulem}
\usepackage{url}
\usepackage{xcolor}
\usepackage{cite}
\usepackage{graphics}

\title{In Defense of Kalman Filtering for Polyp Tracking from Colonoscopy Videos}
\name{David Butler$^{1}$  \quad Yuan Zhang$^{1}$  \quad Tim Chen$^{1}$ \quad Seon Ho Shin$^{2}$ \quad Rajvinder Singh$^{2}$  \quad Gustavo Carneiro$^{1}$}
\address{$^{1}$ School of Computer Science, University of Adelaide \\ $^{2}$ Faculty of Health and Medical Sciences, University of Adelaide }

\begin{document}


\maketitle

\begin{abstract}
Real-time and robust automatic detection of polyps from colonoscopy videos are essential tasks to help improve the performance of doctors during this exam.
The current focus of the field is on the development of accurate but inefficient detectors that will not enable a real-time application.
We advocate that the field should instead focus on the development of simple and efficient detectors that can be combined with effective trackers to allow the implementation of real-time polyp detectors.
In this paper, we propose a Kalman filtering tracker that can work together with  powerful, but efficient detectors, enabling the implementation of real-time polyp detectors.
In particular, we show that the combination of our Kalman filtering with the detector PP-YOLO shows state-of-the-art (SOTA) detection accuracy and real-time processing.  More specifically, our approach has SOTA results on the CVC-ClinicDB dataset, with a recall of 0.740, precision of 0.869, $F_1$ score of 0.799, an average precision (AP) of 0.837, and can run in real time (i.e., 30 frames per second). 
We also evaluate our method on a subset of the Hyper-Kvasir annotated by our clinical collaborators, resulting in 
SOTA results, with a recall of 0.956, precision of 0.875, $F_1$ score of 0.914, AP of 0.952, and can run in real time\footnote{Supported by Australian Research Council through grants DP180103232 and FT190100525.}.

\end{abstract}


\vspace{-.1in}
\section{Introduction}
\vspace{-.1in}

Colorectal cancer is one of the most common forms of cancer~\cite{https://doi.org/10.3322/caac.21395}. The early detection and removal of colorectal polyps before they become malignant is known to improve long-term outcomes~\cite{ZorronChengTaoPu2019}. 
However, the effectiveness of a clinician to detect polyps during a colonoscopy can be affected by many factors, such as experience, professional background, time of day, and length of the procedure, leading to a miss rate that has been measured to be as high as $22\%-28\%$ \cite{Leufkens24.04.2012}.
The performance of clinicians can be improved with the use of a real-time system that can automatically detect polyps during the colonoscopy exam.


The real-time detection of colorectal polyps from colonoscopy videos is an application that has received much attention in recent years \cite{Lee2020, aipdt, Podlasek22.04.2021}.
Real-time object detection has not been possible without considerable sacrifices in accuracy until recently with the innovation of architectures such as YOLO (You Only Look Once) \cite{redmon2016look}, which shows competitive, but generally worse detection accuracy than its more complex counterparts.
The use of object trackers is a natural way to compensate for the relatively poorer accuracy of these more efficient detectors, but the field has not focused too much on this topic, as evidenced by the small number of publications~\cite{7840040, NOGUEIRARODRIGUEZ2021721}.

\begin{figure}
\centering
	\includegraphics[height=0.8in, width=0.8in]{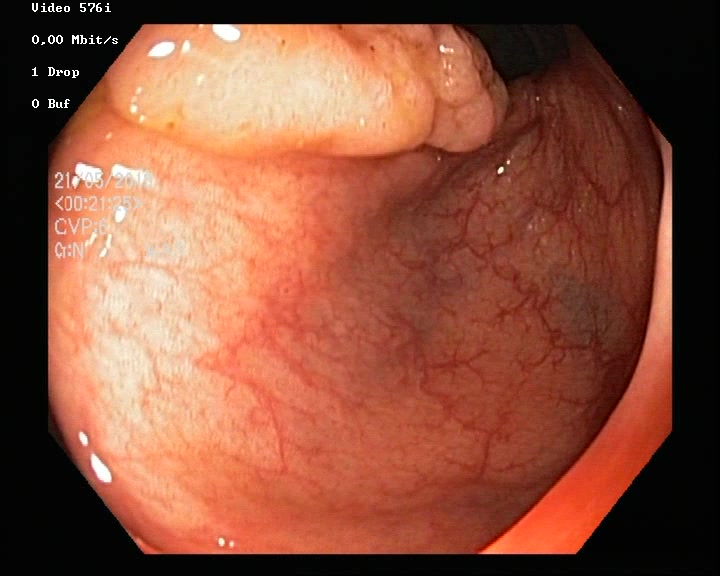}
	\includegraphics[height=0.8in, width=0.8in]{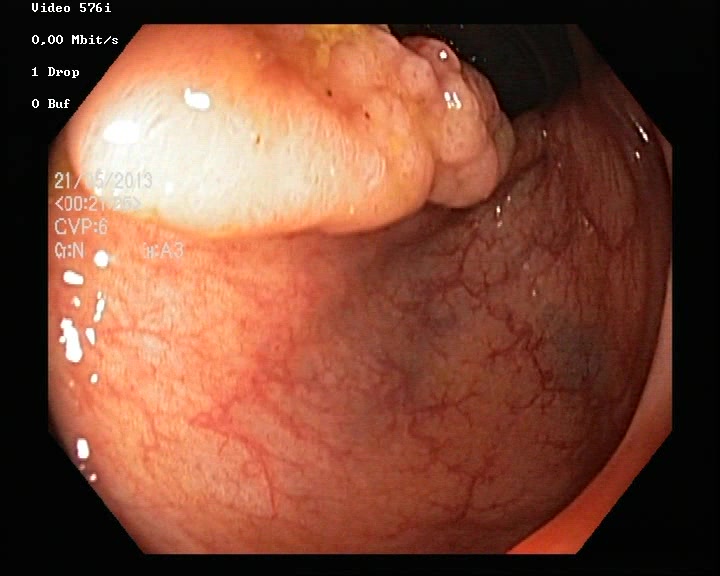}
	\includegraphics[height=0.8in, width=0.8in]{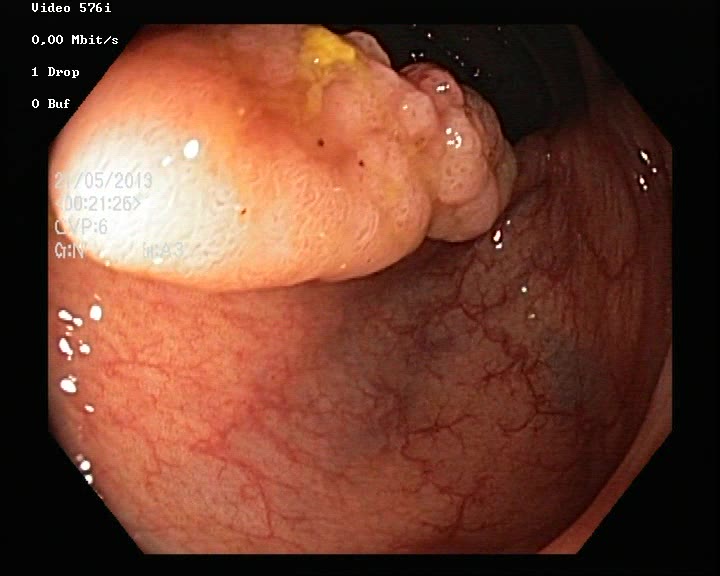}
	\caption{Polyp in video (5 frames apart)}
	\vspace{-15pt}
\end{figure}

This may be due to the lack of publicly available datasets with fully annotated video sequences. 
Even with such constraint, we advocate that the field must focus on the development of simple object trackers that, when combined with efficient detectors, such as YOLO, can produce real-time robust polyp detectors.


In this paper, we propose a simple and effective colorectal polyp tracker based on Kalman filtering that works with relatively accurate but efficient detectors.
In particular, we combine our proposed Kalman filtering with the PP-YOLO detector~\cite{long2020ppyolo}.
We show that our method has state-of-the-art (SOTA) results on the CVC-ClinicDB dataset~\cite{pmid25863519}, with a recall of 0.740, precision of 0.869, $F_1$ score of 0.799, an average precision (AP) of 0.837, and can run at a frame rate of 31.6 frames per second (fps).
We also evaluate our method on a subset of the Hyper-Kvasir~\cite{Borgli2020} annotated by our clinical collaborators, resulting in 
SOTA results, with a recall of 0.956, precision of 0.875, $F_1$ score of 0.914, an AP of 0.952, and can run at 31.6 fps.





\vspace{-.1in}
\section{Related Work}
\vspace{-.1in}

\subsection{Object Detection}
\vspace{-.05in}

The current state-of-the-art methods for object detection all rely on CNNs and can be classified as either two-stage or one-stage detectors \cite{Xiao2020}. Two-stage detectors such as Regional Convolution Neural Networks (R-CNN)~\cite {girshick2014rich} first propose regions of interest (RoIs) that may contain an object, then each is classified separately by a CNN. In contrast, one-step detectors such as YOLO \cite{redmon2016look}, SSD (Single Shot Detector) \cite{Liu_2016}, and RetinaNet \cite{lin2018focal} predict both the bounding boxes and the classes in the same step. 

\vspace{-.05in}
\subsection{Object Tracking}
\vspace{-.05in}

Object tracking is the task of tracking unique objects through a sequence of video frames. Deep learning methods for this task can be categorised as feature extraction-based or end-to-end methods \cite{8964761}. Feature extraction-based methods separate the detection and tracking tasks \cite{8964761}. Detection is done with object detection networks such as R-CNN \cite{girshick2014rich}, SDD \cite{Liu_2016}, or YOLO \cite{redmon2016look} and the extracted features and/or bounding boxes are used for tracking with either classical methods such as Kernel Correlation Filters \cite {Bu2016MultipleOT} or recurrent neural networks \cite{Ghaemmaghami2017TrackingOH}.
Alternatively, end-to-end methods combine both detection and tracking into a single step. These include Siamese networks, patch networks, and graph-based networks \cite{8964761}. Methods that use deep learning for tracking learn both spatial and temporal patterns, and are therefore currently only suited to domains with available annotated frame sequences. Additionally, due to their increased complexity, they tend to be slower than other methods \cite{aipdt}. 

\vspace{-.05in}
\subsection{Polyp Detection}
\vspace{-.05in}

Competitive polyp detection methods are mainly based on deep learning methods (e.g., Faster R-CNN and SSD)~\cite{9098663, Lee2020}. As there is a trade-off between accuracy and speed, many methods with SOTA accuracy are incapable of real-time detection \cite{9098663}, resulting in limited clinical applicability. Hence, the field has focused on the development of real-time detectors that retain some of the accuracy of the SOTA approaches~\cite{long2020ppyolo} 

Such relatively poorer accuracy can be compensated by the use of temporal information, such as a voting window \cite{Lee2020} or a decision tree \cite{10.1007/978-3-319-67543-5_3} that works over successive frames to reduce the number of false positives. Even though such methods are promising, they were evaluated with older and inaccurate object detectors. 
SOTA detection methods that harness temporal information have been proposed \cite{9098663}, but require a large training set containing fully annotated video sequences, which is challenging to acquire.





Recently proposed polyp tracking methods can be categorised by whether they use online or offline learning. Online methods consist of a trained object detector that provides features, bounding boxes, or image patches to a tracker that learns the representation of a specific polyp at inference time. Such methods include RYCO \cite{ZHANG2018209} and AIPDT \cite{aipdt}, both of which use discriminate correlation filter-based trackers that learn image patches. While AIPDT can track polyps in real time, online methods may have difficulty tracking polyps that rapidly change appearance due to alterations in lighting or occluding material, as they learn a narrow representation during inference. 
Alternatively, offline methods can learn temporal patterns from large data sets. A recent example of an offline method is the Spatial-Temporal Feature Transform (STFT) \cite{stft} that uses the current frame and one or more previous frames as inputs. STFT produces SOTA detection accuracy, but is unable to run in real-time.


A major challenge in the field is the ability to provide a fair comparison between these methods because some methods have only been evaluated on private datasets \cite{AHMAD2019AB647, 9098663}. To mitigate this problem, we use two public datasets~\cite{pmid25863519,Borgli2020} to compare our approaches to SOTA methods~\cite{lin2018focal,stft,long2020ppyolo}.

\vspace{-.1in}
\section{Method}
\vspace{-.1in}

Our training set is denoted by $\mathcal{D}=\{(\mathbf{x}_i,\mathbf{y}_i)\}_{i=1}^{|\mathcal{D}|}$ containing images $\mathbf{x} \in \mathcal{X} \subset \mathbb{R}^{H \times W \times C}$ of size $H \times W$ with $C$ colour channels, and labels $\mathbf{y} \in \mathcal{Y} \subset \mathbb{R}^4$ consisting of the bounding box of the polyp (i.e., 2-dimensional centre, and width and height).  This training set is used to train a polyp detector (e.g., PP-YOLO~\cite{long2020ppyolo} or RetinaNet~\cite{lin2018focal}), denoted by a $\theta$-parameterised function $f_{\theta}:\mathcal{X} \to \mathcal{B}$ 
that takes an image $\mathbf{x} \in \mathcal{X}$ and produces a set of bounding boxes $\mathcal{B} = \{ ( \mathbf{y}_b,c_b ) \}_{b=1}^{|\mathcal{B}|}$, with $c_b \in [0,1]$
denoting the probability that the model is confident in the detection. 
The testing set is represented by a set of $|\mathcal{T}|$ colonoscopy videos containing $T$ frames, denoted by $\mathcal{T} = \{ (\mathbf{x}_{i,t},\mathbf{y}_{i,t}) \}_{i=1,t=1}^{|\mathcal{T}|,T}$. An important difference of the testing set, compared with the training set, is that video frames may not have any polyp, so the annotation $\mathbf{y}_{i,t} \in \mathcal{Y}$ is redefined to also contain $\emptyset$.

Our main contribution is the development of a simple and effective Kalman filtering that runs during testing to improve the accuracy of the detector. We describe our Kalman filtering in more detail below.

\vspace{-.05in}
\subsection{Kalman Filtering}
\vspace{-.05in}

Kalman filtering is a method for estimating unknown variables in a discrete time linear dynamics system using measurements that are assumed to have Gaussian noise \cite{5512258}. We use Kalman filtering to estimate a polyp location and size across video frames by using the output of the detector as measurements. The state vector for frame $t\in\{1,...,T\}$ is defined as:
\begin{equation}
    \mathbf{s}_t = [x_t, y_t, w_t, h_t, \Delta x_{t}, \Delta y_{t}, \Delta w_{t}, \Delta h_{t}]^T,
\end{equation}
where $\mathbf{y}_t = [x_t, y_t, w_t, h_t]$ ($x_t, y_t$ denote the bounding box centre, $w_t, h_t$ represent the width and height of the bounding box), and $\Delta x_{t}, \Delta y_{t}, \Delta w_{t}, \Delta h_{t}$ are the rates of change of the bounding box centre and size, which are estimated using the relationship described by the state transition model $\mathbf{F}$ below.


Kalman filtering assumes the system transitions from time step $t - 1$ to $t$ with noise $\mathbf{w_t} \sim \mathcal{N}(\mathbf{0}, \mathbf{Q})$ (where $\mathcal{N}(\mathbf{0}, \mathbf{Q})$ denotes the Gaussian distribution of mean $\mathbf{0}$ and covariance $\mathbf{Q}$) according to the equation:
\begin{equation}
    \mathbf{s}_t = \mathbf{F} \mathbf{s}_{t-1} + \mathbf{w}_t,
\end{equation}
where $\mathbf{F}$ denotes the state transition model describing how the linear dynamic changes through time, defined as
\begin{equation*}
    \mathbf{F} = 
    \begin{bmatrix}
    1 & 0 & 0 & 0 & \Delta t & 0 & 0 & 0 \\
    0 & 1 & 0 & 0 & 0 & \Delta t & 0 & 0 \\
    0 & 0 & 1 & 0 & 0 & 0 & \Delta t & 0 \\
    0 & 0 & 0 & 1 & 0 & 0 & 0 & \Delta t \\
    0 & 0 & 0 & 0 & 1 & 0 & 0 & 0 \\
    0 & 0 & 0 & 0 & 0 & 1 & 0 & 0 \\
    0 & 0 & 0 & 0 & 0 & 0 & 1 & 0 \\
    0 & 0 & 0 & 0 & 0 & 0 & 0 & 1 \\
    \end{bmatrix}.
\end{equation*}
The relation between the system state $\mathbf{s}_t$ and measurement $\mathbf{z}_t$ with noise $\mathbf{v}_t \sim \mathcal{N}(\mathbf{0}, \mathbf{R})$ is computed as:
\begin{equation*}
    \mathbf{z}_t = \mathbf{H} \mathbf{s}_t + \mathbf{v}_t,
\end{equation*}
where $\mathbf{R}$ represents the noise covariance matrix for the measurements, and $\mathbf{H}$ describes the measurement model, defined as:
\begin{equation*}
    \mathbf{H} = 
    \begin{bmatrix}
    \mathbf{I}_{4 \times 4} & \mathbf{0}_{4 \times 4}
    \end{bmatrix},
\end{equation*}
with $\mathbf{I}_{4 \times 4}$ denoting a $4 \times 4$ identity matrix, and $\mathbf{0}_{4 \times 4}$ a $4 \times 4$ zero matrix.
The filtering process consists of two steps: a prediction step and an update step. The prediction step estimates the \emph{a priori} state $\mathbf{\hat{s}'_{t}}$ and error $\mathbf{\hat{P}'_{t}}$:
\begin{align}
\begin{split}
    k_{predict} (\mathbf{\hat{s}}_{t-1}) = \mathbf{\hat{s}'}_{t} &= \mathbf{F} \mathbf{\hat{s}}_{t-1},\\
    \mathbf{\hat{P}'}_{t} &= \mathbf{F} \mathbf{\hat{P}}_{t-1} \mathbf{F}^T + \mathbf{Q},
\end{split}
\label{eq:kalman_predict}
\end{align}
where $\mathbf{\hat{s}'}_{t}$ is the \emph{a priori} state estimate, $\mathbf{\hat{P}'}_{t}$ denotes the predicted \emph{a priori} estimate covariance, and $\mathbf{\hat{P}}_{t-1}$ represents \emph{a posteriori} estimated covariance from step $t-1$.
The update step uses the best matching measurement from the detector, 
$\mathbf{z}_t = \mathbf{y}_b$, 
as determined by data association (\ref{eq:overlap_score}) to estimate the \emph{a posteriori} state $\mathbf{\hat{s}}_{t}$ and error $\mathbf{\hat{P}'}_{t}$. Before this is done, the Kalman gain, $\mathbf{K_t}$, is calculated and used to scale the effect of the measurement given its estimated accuracy:
\begin{align}
\begin{split}
    \mathbf{K_t} &= \mathbf{\hat{P}'}_{t} \mathbf{H}^T(\mathbf{H} \mathbf{\hat{P}'}_{t} \mathbf{H}^T + \mathbf{R})^{-1},\\
    \mathbf{\hat{P}_t} &= (\mathbf{I} - \mathbf{K_t} \mathbf{H}) \mathbf{\hat{P}'}_{t},\\
    \mathbf{\hat{s}}_{t} &= \mathbf{\hat{s}'}_{t} + \mathbf{K_t} (\mathbf{z}_t - \mathbf{H} \mathbf{\hat{s}'}_{t}).
\end{split}
\label{eq:kalman_update}
\end{align}

The noise covariances $\mathbf{Q}$ and $\mathbf{R}$ are difficult to estimate, so following~\cite{5512258}, we assume that noise in state transition model is independent, with $\mathbf{Q} = \mathbf{I}_{8 \times 8} \times 0.01$, where $\mathbf{I}_{8 \times 8}$ is an $8 \times 8$ identity matrix; and $\mathbf{R} = \mathbf{0}_{4 \times 4}$ (i.e., a $4 \times 4$ zero matrix).


\vspace{-.05in}
\subsection{Data Association}
\vspace{-.05in}

At each image of the video, there can be multiple polyp detections, so it is important that tracked polyps are not confused with one another. Hence, our method also relies on data association to enable multi-target tracking.
Data association links detected bounding boxes (from the  detector) with tracked bounding boxes (from the Kalman filter).
Assuming that the detections from $t^{th}$ frame produces $\mathcal{B} = f_{\theta}(\mathbf{x}_t)$, where $\mathcal{B} = \{ \mathbf{y}_b,c_b  \}_{b=1}^{|\mathcal{B}|}$, we first run a non-max suppression, where we only keep the highest confidence detections among bounding boxes that have an overlap difference score larger than $\alpha$.  This  score between bounding boxes $\mathbf{y}_i$ and $\mathbf{y}_j$ is computed with:
\begin{equation}
    s(\mathbf{y}_i, \mathbf{y}_j) = \sqrt{\lvert 4 w_i h_i - 4 w_i h_j \rvert} + \sqrt{(x_i - x_j)^2 + (y_i - y_j)^2},
    \label{eq:overlap_score}
\end{equation}
where $(x_i,y_i)$ denotes the bounding box centre and $(w_i,h_i)$ represents the bounding box width and height (similarly for $(x_j,y_j,w_j,h_j)$).
The non-max suppressed detections are stored in the set $\tilde{\mathcal{B}}_{t}$.
Then, we use Kalman prediction from~\eqref{eq:kalman_predict} to estimate the location of tracked polyps from the previous frame $\mathcal{P}_{t-1} = \{ \mathbf{s}_p,c_p  \}_{p=1}^{|\mathcal{P}_{t-1}|}$, as in 
$\tilde{\mathcal{P}}_{t} = \{\{k_{predict}(\mathbf{s}_p), c_p\} : \forall \{\mathbf{s}_p,c_p\} \in \mathcal{P}_{t-1}\}$
Next, we associate the detections from $\tilde{\mathcal{B}}_{t}$ with polyps in $\tilde{\mathcal{P}}_{t}$ by matching detections with overlap difference score~\eqref{eq:overlap_score} smaller than $\beta$ and detection confidence larger than $\phi$. These matched detections are then used in the Kalman update in~\eqref{eq:kalman_update} to update the list of tracked polyps in $\tilde{\mathcal{P}}_{t}$.  
If a tracked polyp has not been updated within $\epsilon$ time steps, it is discarded.
$\mathcal{P}_{t}$ then consists of the updated polyps, as well as unmatched detections from $\tilde{\mathcal{B}}_{t}$ with confidence larger than $\gamma$.

\vspace{-.1in}
\section{Experimental Setup}
\vspace{-.1in}

\subsection{Datasets}
\vspace{-.05in}

We use two public datasets to test our system: Hyper-Kvasir and CVC-ClinicDB datasets.
The Hyper-Kvasir \cite{Borgli2020} dataset is a broad gastrointestinal endoscopy dataset with examples of many different pathological findings. It has 1000 colonoscopy images of colorectal polyps annotated with bounding boxes, as well as 72 colonoscopy videos that contain colorectal polyps. These videos are not labelled frame-by-frame, so we select six to be annotated with the supervision of a gastroenterologist. The videos are selected so there is no overlap with the 1000 annotated images. This is done by using a MobileNet \cite{howard2017mobilenets} model trained on the ImageNet dataset \cite{deng2009imagenet} and then calculating the cosine similarity between the features of each image-frame pair. This method is validated on known matches between images and videos. 

Two separate configurations are used for training, the first is image-based, which uses 800 of the Hyper-Kvasir images for training, and the remaining 200 for validation. The second is video-based and uses 4 of the Hyper-Kvasir videos for training.

Testing is done with the 2 remaining Hyper-Kvasir videos and the CVC-ClinicDB dataset~\cite{pmid25863519}, which is composed of 612 annotated colorectal polyp images. While CVC-ClinicDB is not strictly a video database, it comprised of many short colonoscopy sequences, which allows for methods like Kalman filtering and STFT~\cite{stft} to take advantage of temporal information.

\vspace{-.05in}
\subsection{Detectors and Training}
\vspace{-.05in}

We test the application of our Kalman filtering to three different types of polyp detectors: RetinaNet~\cite{lin2018focal}, 
PP-YOLO~\cite{long2020ppyolo} and STFT~\cite{stft}. While two versions of RetinaNet and PP-YOLO are trained exclusively on either the image or video data sets, STFT can only be trained with the video data set.
PP-YOLO, RetinaNet and STFT are trained for 60,000, 8,000 and 35,000 iterations respectively, with a batch sizes of 5\footnote{The authors of PP-YOLO use a batch size of 24, but due to hardware limitations, we were restricted to 5, which may cause the model to not converge optimally.}, 4 and 1, and learning rates of 0.000625, 0.0001 and 0.000125. All methods are optimised with SGD with a momentum factor of 0.9. 
The 4 Hyper-Kvasir videos in the training set are used to estimate the parameters of the data association for Kalman filtering, where average precision (AP -- explained below in Sec.~\ref{sec:evaluation}) is used to select the following parameter values for PP-YOLO and STFT: $\alpha = 700$, $\beta = 295$, $\epsilon = 17$, $\phi = 0.17$, and $\gamma = 0.0575$.
For RetinaNet we instead use $\alpha = 700$, $\beta = 295$, $\epsilon = 1$, $\phi = 0.8$, and $\gamma = 0.09$.




\vspace{-.05in}
\subsection{Evaluation}
\label{sec:evaluation}
\vspace{-.05in}

The detection methods are quantitatively assessed with precision, measured with $P = \frac{TP}{TP+FP}$, and recall $R=\frac{TP}{TP+FN}$. 
True positives are determined by the intersection-over-union (IoU); the area of the intersection of the predicted bounding box and the ground truth bounding box divided by the area of their union. For this application an IoU $\geq 0.2$ is considered to be a true positive.
We also evaluate the detectors with average precision (AP), which is the average value of the precision as a function of the recall in the interval $[0, 1]$. It is defined with $P_n$ and $R_n$ being the precision and recall at the $n^{th}$ threshold as \cite{article}:
\begin{align*}
    AP = \sum_n (R_n - R_{n - 1}) P_n.
\end{align*}
We also calculate the $F_1$ score; the harmonic mean of the precision and recall, defined by $F_1 = 2 \times \frac{P \times R}{P + R}$.
Additionally, the Free-response receiver operating characteristic (FROC) curve is displayed by plotting recall against the number of false positives per minute of video (assuming 25 frames per second) as the classification threshold is changed~\cite{Bandos2009}.
To assess the efficiency of the methods we report the amount of time they need to process a single image and the number of frames per second (fps) they can process.

The performance of the system is evaluated on the entire CVC-ClinicDB dataset and the 2 annotated test videos from the Hyper-Kvasir dataset. The computer used in this evaluation was equipped with Intel i7-6700K, 16GB memory, and Nvidia GTX 1070 8GB.

\vspace{-.1in}
\section{Results and Discussion}
\vspace{-.1in}

\begin{table}
  \caption{Results on CVC-ClinicDB}
  \resizebox{1\columnwidth}{!}{%
  \begin{tabular}{ccccccc}
    \toprule
    Method & Precision & Recall & $F_1$ & $AP$ & Inference Time (SD) & Frame Rate \\
    \midrule
    \multicolumn{7}{c}{Image Trained}\\
    \hline
    RetinaNet & 0.836 & 0.897 & 0.865 & 0.942 & 113.03 (5.43) & 8.8 \\
    RetinaNet + Kalman & 0.900 & 0.936 & 0.918 & $\mathbf{0.977}$ & 114.74 (5.96) & 8.7 \\
    PP-YOLO & 0.869 & 0.873 & 0.871 & 0.904 & $\mathbf{30.58}$ (0.19) & $\mathbf{32.7}$ \\
    PP-YOLO + Kalman & $\mathbf{0.948}$ & $\mathbf{0.948}$ & $\mathbf{0.948}$ & 0.963 & 31.60 (0.53) & 31.6 \\
    \hline
    \multicolumn{7}{c}{Video Trained}\\
    \hline
    RetinaNet & 0.390 & 0.600 & 0.294 & 0.189 & 113.03 (5.43) & 8.8 \\
    RetinaNet + Kalman & 0.390 & 0.600 & 0.473 & 0.453 & 114.74 (5.96) & 8.7 \\
    STFT & 0.509 & 0.608 & 0.554 & 0.545 & 529.10 (33.10) & 1.89 \\
    STFT + Kalman & 0.737 & 0.796 & 0.765 & $\mathbf{0.868}$ & 531.54 (34.75) & 1.88 \\
    PP-YOLO & 0.467 & $\mathbf{0.827}$ & 0.597 & 0.654 & $\mathbf{30.58 (0.19)}$ & $\mathbf{32.7}$ \\
    PP-YOLO + Kalman & $\mathbf{0.869}$ & 0.740 & $\mathbf{0.799}$ & 0.837 & 31.60 (0.53) & 31.6 \\
    \bottomrule
  \end{tabular}%
  }
  \label{table:cvc}
\end{table}

\begin{figure}[t]
    \begin{center}
	\includegraphics[width=2.7in]{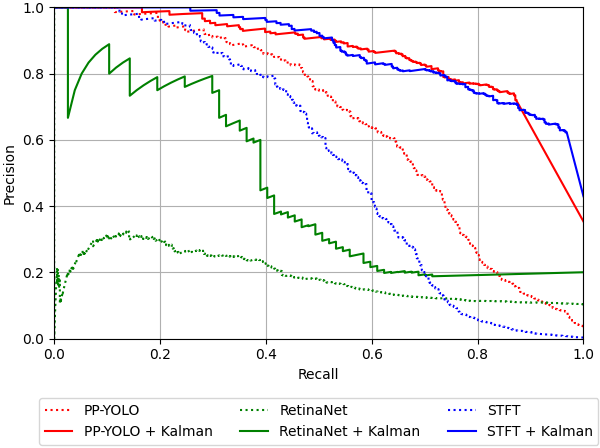}
	\end{center}
	\caption{Precision-recall curves of video trained methods on CVC-ClinicDB}
	\label{fig:prc_cvc}
\end{figure}
\begin{figure}[t]
    \begin{center}
	\includegraphics[width=2.7in]{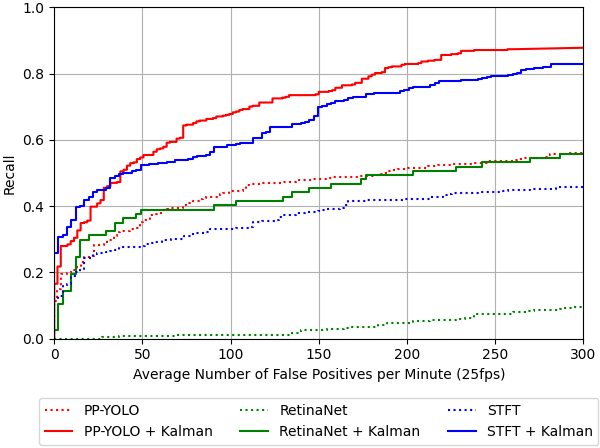}
	\end{center}
	\caption{FROC curves of video trained methods on CVC-ClinicDB}
	\label{fig:froc_cvc}
\end{figure}

Table~\ref{table:cvc} shows the precision, recall, $F_1$ and $AP$ improvements brought by the application of Kalman filtering to the RetinaNet, STFT and PP-YOLO detectors on the CVC-ClinicDB dataset. 
It can be seen that for almost all cases, the use of our Kalman filtering improves precision, recall, $F_1$ and $AP$ for all detectors with negligible computational cost.
The prevision and recall improvement on CVC-ClinicDB can be seen in Fig.~\ref{fig:prc_cvc} for all recall values, and the FROC curve in Fig.~\ref{fig:froc_cvc} shows the clear advantage of using Kalman filtering for all detectors.

\begin{table}
  \caption{Results On Hyper-Kvasir Videos}
  \resizebox{1\columnwidth}{!}{%
  \begin{tabular}{ccccccc}
    \toprule
    Method & Precision & Recall & $F_1$ & $AP$ & Inference Time (SD) & Frame Rate \\
    \midrule
    \multicolumn{7}{c}{Image Trained}\\
    \hline
    RetinaNet & 0.839 & 0.907 & 0.872 & 0.919 & 113.03 (5.43) & 8.8 \\
    RetinaNet + Kalman & 0.840 & $\mathbf{0.969}$ & 0.900 & $\mathbf{0.965}$ & 114.74 (5.96) & 8.7 \\
    PP-YOLO & 0.870 & 0.950 & 0.908 & 0.936 & $\mathbf{30.58 (0.19)}$ & $\mathbf{32.7}$ \\
    PP-YOLO + Kalman & $\mathbf{0.875}$ & 0.956 & $\mathbf{0.914}$ & 0.952 & 31.61 (0.53) & 31.6 \\
    \hline
    \multicolumn{7}{c}{Video Trained}\\
    \hline
    RetinaNet & 0.326 & 0.462 & 0.382 & 0.310 & 113.03 (5.43) & 8.8 \\
    RetinaNet + Kalman & 0.626 & 0.419 & 0.502 & 0.421 & 114.74 (5.96) & 8.7 \\
    STFT & 0.409 & 0.920 & 0.566 & 0.474 & 529.10 (33.10) & 1.89 \\
    STFT + Kalman & $\mathbf{0.685}$ & 0.918 & $\mathbf{0.785}$ & $\mathbf{0.787}$ &  531.54 (34.75) & 1.88 \\
    PP-YOLO & 0.269 & 0.210 & 0.236 & 0.137 & $\mathbf{30.58 (0.19)}$ & $\mathbf{32.7}$ \\
    PP-YOLO + Kalman & 0.608 & 0.236 & 0.340 & 0.196 & 31.60 (0.53) & 31.6 \\
    \bottomrule
  \end{tabular}%
  }
  \label{table:hk}
\end{table}
\begin{figure}[t!]
\begin{center}
	\includegraphics[width=2.7in]{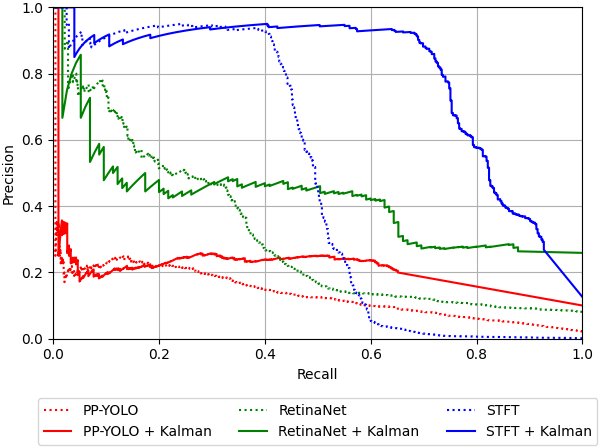}
	\end{center}
	\caption{Precision-recall curves of video trained methods on Hyper-Kvasir}
	\label{fig:prc_hk}
\end{figure}
\begin{figure}
\begin{center}
	\includegraphics[width=2.7in]{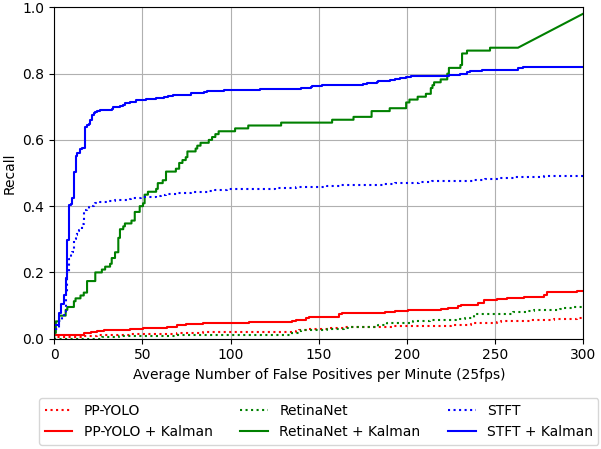}
	\end{center}
	\caption{FROC curves of video trained methods on Hyper-Kvasir}
	\label{fig:froc_hk}
\end{figure}

Similarly, Table~\ref{table:hk} shows the results of RetinaNet, STFT and PP-YOLO (with and without Kalman filtering) on Hyper-Kvasir.
Again, the computational cost of Kalman filtering is negligible, where PP-YOLO with and without Kalman filtering is the only approach with real-time processing. 
The prevision and recall improvement on Hyper-Kvasir is displayed in Fig.~\ref{fig:prc_hk}, and the FROC curve in Fig.~\ref{fig:froc_hk} shows that Kalman filtering helps to increase the recall for all values of false positives. Note that for all methods, the only approaches that can produce real-time analysis is PP-YOLO with and without Kalman filtering.

\vspace{-.1in}
\section{Conclusion}
\vspace{-.1in}

For detecting and tracking colorectal polyps, Kalman filtering always improves detection accuracy with negligible additional computational cost.
The difference in performance on CVC-ClinicDB and Hyper-Kvasir shows the need for public datasets that better represent realistic operating conditions for colorectal polyp detection systems. To show that these systems perform well enough to be widely deployed, they will need to be evaluated on datasets that are representative of their deployed environment. 

\bibliographystyle{IEEEbib}
\bibliography{strings,refs,sample-base}
\end{document}